\documentclass{cta-author}
\newcommand{\sthanks}[1]{\thanks{#1}}
\usepackage{stfloats}
\usepackage{pifont}
\usepackage{float}
\usepackage{amssymb}
\usepackage[dvipsnames]{xcolor}
\usepackage{subcaption}
\usepackage{placeins}
\usepackage{multirow}
\usepackage{booktabs}
\usepackage{graphicx}
\usepackage{booktabs}       
\usepackage{multirow}

\newcommand\jjj[1]{\textcolor{black}{#1}}

{}
{}
{}

\begin{document}

\supertitle{Submission Template for IET Research Journal Papers}

\title{RoboLoc: A Benchmark Dataset for Point Place Recognition and Localization in Indoor-Outdoor Integrated Environments}


\author{
\au{Jaejin Jeon$^{1 \dag }$},
\au{Seonghoon Ryoo$^{1 \dag}$}, 
\au{Sang-Duck  Lee$^{2}$},  
\au{Soomok Lee$^{3\corr}$ and 
\au{Seungwoo Jeong$^{2\corr}$} }
}

\address{
\add{1}{Department of Data, Network and AI, Ajou University, Ajou University, 206, World cup-ro,  Suwon-si, S.Korea }
\add{2}{ Korea Railroad Research Institute, 176 Cheoldomuseum-ro, Uiwang-si, S.Korea}
\add{3}{ Department of Mobility Engineering, Ajou University, 206 World cup-ro,  Suwon-si,  S.Korea} 
\sthanks{$^\dagger$ Equal contribution}\\
\email{swjeong@krri.re.kr and soomoklee@ajou.ac.kr}; Authors with * are co-corresponding authors}


\begin{abstract}
Robust place recognition is essential for reliable localization in robotics, particularly in complex environments with frequent indoor-outdoor transitions. However, existing LiDAR-based datasets often focus on outdoor scenarios and lack seamless domain shifts. In this paper, we propose RoboLoc, a benchmark dataset designed for GPS-free place recognition in indoor-outdoor environments with floor transitions. RoboLoc features real-world robot trajectories, diverse elevation profiles, and transitions between structured indoor and unstructured outdoor domains. We benchmark a variety of state-of-the-art models, point-based, voxel-based, and BEV-based architectures, highlighting their generalizability domain shifts. RoboLoc provides a realistic testbed for developing multi-domain localization systems in robotics and autonomous navigation.
\end{abstract}
\maketitle

\section{Introduction}

Understanding pose and location has been one of the main challenges in robotics. Traditionally, most approaches have relied on vision-based place recognition methods. Visual place recognition, also known as loop closure detection in simultaneous localization and mapping (SLAM), involves identifying a matching image from a database given a query image~\cite{arandjelovic2016netvlad}.

With the recent adoption of range measurement sensors, particularly LiDAR, place recognition has significantly advanced in terms of robustness and accuracy. LiDAR is less sensitive to illumination changes and provides high-precision distance measurements, making it a strong modality for mapping and localization. Accordingly, several studies have developed loop closure detection techniques based on point cloud descriptors~\cite{uy2018pointnetvlad, kim2018scan, ma2022overlaptransformer}.

A number of benchmark datasets have played a central role in this progress. For example, Oxford RobotCar~\cite{barnes2020oxford}, KITTI~\cite{geiger2012we}, and MulRan~\cite{kim2020mulran} provide long-term traversals in urban driving scenarios under diverse conditions, while PointNetVLAD~\cite{uy2018pointnetvlad} introduced curated LiDAR-only datasets highlighting viewpoint variations and loop closure cases. More recent efforts, such as HeLiPR~\cite{jung2024helipr} and MCD~\cite{nguyen2024mcd}, extend coverage to heterogeneous sensors and crowded outdoor environments. Collectively, these datasets have advanced LiDAR-based localization research. However, they remain heavily biased toward outdoor driving, lacking comprehensive indoor sequences and, more critically, seamless transitions across indoor and outdoor domains. This limitation hinders progress for service robots and autonomous mobile platforms that routinely operate in GPS-denied and mixed environments such as campuses, building complexes, or underground facilities. \jjj{Recent advances in 3D perception, including transformer-based geometry modeling~\cite{tong2024edge} and neural rendering-based stereo~\cite{tong2025neural}, further underscore the need for benchmarks that capture domain transitions and structural diversity.
}

\jjj{To address this gap, we introduce the \textbf{RoboLoc} dataset, a new benchmark explicitly designed for multi domain place recognition and loop closure. 
Unlike prior datasets that remain biased toward outdoor driving or lack seamless domain transitions, RoboLoc for the first time provides continuous trajectories that naturally span indoor and outdoor environments. 
The dataset was collected using a LiDAR-equipped mobile robot across building interiors, hallways, entrances, courtyards, and open roads, thereby capturing domain boundaries that existing benchmarks overlook. 
In addition, RoboLoc uniquely incorporates diverse elevation profiles, ranging from flat terrain to steep mountainous slopes, and supports navigation across multi level indoor floors. 
This combination of indoor–outdoor transitions, vertical complexity, and structural diversity reflects realistic challenges encountered by service robots and autonomous platforms in GPS-denied campuses and building complexes. 
By explicitly addressing these missing aspects, RoboLoc offers a unique and challenging testbed for advancing localization research in complex, mixed domain environments.}

Our main contributions are summarized as follows:
\begin{itemize}
    \item We present a novel dataset that bridges indoor and outdoor environments, providing unified benchmarking for place recognition in realistic service robot and autonomous mobile robot (AMR) scenarios where domain transitions are common.
    \item We benchmark state-of-the-art place recognition models on RoboLoc, demonstrating its challenging constraints and highlighting the generalization requirements posed by mixed-domain conditions.
    \item We emphasize challenging scenarios with varying elevation profiles and multi level indoor environments, enabling in-depth analysis of algorithm strengths and weaknesses in realistic 3D settings.
\end{itemize}

The remainder of this paper is organized as follows: Section~2 reviews related works on LiDAR-based place recognition and datasets. Section~3 describes the RoboLoc dataset, sensing platform, and collection methodology. Section~4 presents experimental evaluations and benchmarking results. Section~5 concludes with a discussion on limitations and future directions.
\newcommand{\cmark}{\ding{51}}      
\newcommand{\sone}{\ding{72}}       
\newcommand{\stwo}{\sone\sone}      
\newcommand{\sthree}{\sone\sone\sone} 

\begin{table*}[!t]
\centering
\small
\footnotesize
\caption{Dataset comparison with respect to range sensor configuration and suitability for place-recognition study}
\renewcommand{\arraystretch}{1.25}
\setlength{\tabcolsep}{2pt}
\begin{tabular}{llcccccccccc}
\toprule
\multicolumn{2}{c}{\textbf{Datasets}} &
Freiburg &
Ford Campus &
KITTI &
NCLT &
Complex Urban &
nuScenes &
Marulan &
Oxford RobotCar &
MulRan &
\textbf{RoboLoc (ours)} \\
\midrule
\multirow{2}{*}{\textbf{Diversity}} &
Structural &        & \sone & \stwo & \sthree & \sthree & \stwo & \sone & \sone & \sthree & \sthree \\
& Temporal   &        &       & \sone & \sthree & \stwo   & \stwo &       & \sone & \sthree & \stwo \\
\midrule
\multirow{2}{*}{\textbf{Indoor}} & Single floor &  &  &  & \sone &  &  &  & \ &  & \stwo \\
  & Multi floor &  &  &  & &  &  &  & \ &  & \sone \\

\midrule
\multirow{2}{*}{\textbf{Loop}} &
Frequency & \sone & \sone & \sone & \stwo & \stwo & \sone &       & \sone & \sthree & \sone \\
& Reverse   &       &       &       & \sone & \stwo &       &       &       & \sthree & \sone \\
\bottomrule
\end{tabular}
\footnotesize
\end{table*}

\section{Related Work}
\subsection{LiDAR-based 3D Place Recognition}

LiDAR-based place recognition has evolved rapidly from handcrafted descriptors to deep learning-based representations that leverage geometric reasoning and long-range dependencies. Early approaches such as Fast Histogram~\cite{rohling2015fast}, M2DP~\cite{he2016m2dp}, and Scan Context~\cite{kim2018scan} used handcrafted global descriptors derived from 2D projections or statistical features. While computationally efficient, these methods were limited in their ability to generalize across structurally diverse and large-scale environments.

The introduction of learning-based models marked a turning point. PointNetVLAD~\cite{uy2018pointnetvlad} applied PointNet to extract local point features, followed by NetVLAD aggregation to produce global descriptors. LPD-Net~\cite{liu2019lpd} further incorporated handcrafted local features and graph neural networks to improve spatial context modeling. These point-based models, though effective, often struggled with structural preservation and scalability.

To address these limitations, sparse voxel-based architectures gained popularity. MinkLoc3D~\cite{komorowski2021minkloc3d} utilized sparse 3D convolutions with feature pyramids to encode compact yet discriminative descriptors, and its successor MinkLoc3Dv2~\cite{komorowski2022improving} introduced channel attention and deeper layers to enhance robustness to viewpoint and structural variations. CrossLoc3D~\cite{guan2023crossloc3d} and SG-LPR~\cite{jiang2024sg} extended this line of work by incorporating multi-resolution embeddings, semantic features, and alignment across aerial-ground modalities.

Recent developments have shifted toward transformer-enhanced architectures to capture long-range dependencies and improve domain robustness. TransLoc3D~\cite{xu2021transloc3d} augments sparse convolution backbones with adaptive receptive fields and a transformer encoder, while PTC-Net integrates point-wise transformers into voxel encoders to mitigate geometric detail loss during voxelization. OverlapTransformer~\cite{ma2022overlaptransformer}, SphereVLAD++~\cite{zhao2022spherevlad++}, and IS-CAT~\cite{joo2024cat} further explore alternative geometric encodings and cross-modal attention mechanisms to address viewpoint changes and semantic variation.
\jjj{
Beyond place recognition, transformer-based approaches have also been explored in related 3D perception tasks such as depth estimation~\cite{tong2024edge}, reflecting a broader trend toward geometric consistency and domain-robust formulations.
}

Despite these advances, most existing methods are evaluated in either indoor or outdoor only domains, where structural continuity and transition complexity are limited. In practice, mobile robots often navigate across semantically distinct and geometrically inconsistent environments, where changes in viewpoint, occlusion, and domain semantics occur frequently. Under such conditions, especially those involving building entrances, forested paths, or multi-floor transitions, performance can degrade substantially due to perceptual aliasing and sensing artifacts.

To evaluate generalization in such challenging scenarios, datasets like RoboLoc are needed. By embedding diverse environmental types and frequent transitions into continuous trajectories, RoboLoc enables benchmarking of spatial reasoning and domain adaptation in mixed-domain LiDAR-based place recognition.

\subsection{Indoor and Mixed-domain Localization}

Despite advances in LiDAR-based place recognition, most methods are developed and evaluated under domain-isolated conditions. Benchmarks and algorithms often focus on either outdoor urban driving~\cite{geiger2012we, barnes2020oxford, kim2020mulran, nguyen2024mcd} or indoor spaces designed for visual-inertial SLAM. While SG-LPR~\cite{jiang2024sg} and IS-CAT~\cite{joo2024cat} have been evaluated on mixed datasets like NCLT~\cite{carlevaris2016university}, such tests typically exclude full structural transitions, such as those from enclosed indoor corridors to outdoor open spaces.

Recent efforts aim to bridge these structural gaps. DiSCO~\cite{xu2021disco} and LoGG3D-Net~\cite{vidanapathirana2022logg3d} incorporate domain-invariant representations via orientation-aware or hierarchical feature learning. Spatial Relation Graphs~\cite{gong2021two} model scene layout using graph structures over segmented point clouds to enable robust matching across spatial layouts. However, these approaches are typically tested in proxy scenarios such as building lobbies or semi-outdoor spaces, rather than in fully continuous trajectories that span multiple floors and environments.

In parallel, LiDAR-based SLAM frameworks like GLIM~\cite{koide2024glim} achieve robust indoor localization using range-inertial fusion and GPU-accelerated scan matching. Yet, they generally lack global place recognition capabilities. Overall, current methods are not well-suited for realistic mixed-domain navigation where geometric scale, structural topology, and perceptual statistics can shift drastically. This highlights the need for new models and datasets designed for seamless localization across indoor–outdoor transitions in GPS-denied real-world environments.

\subsection{Benchmark Datasets for Place Recognition}

The advancement of place recognition techniques depends heavily on the availability of realistic and structurally diverse datasets. Several widely adopted benchmarks target outdoor environments. For instance, KITTI~\cite{geiger2012we}, Oxford RobotCar~\cite{barnes2020oxford}, and MulRan~\cite{kim2020mulran} provide long-range urban trajectories with varied viewpoints, weather conditions, and seasonal dynamics. These datasets primarily focus on road-based navigation using vehicle-mounted sensors. As a result, they typically lack architectural transitions, pedestrian-scale geometry, or vertical scene complexity.

More recently, datasets like MCD~\cite{nguyen2024mcd} have expanded coverage to pedestrian-rich campus scenes, capturing sequences in non-road environments with dynamic obstacles. However, these datasets are still limited to purely outdoor spaces. They do not include any transitions into building interiors or upward/downward navigation across floors, which are essential for real-world deployment of mobile robots in mixed environments.

Among the few datasets that attempt to include both indoor and outdoor domains, NCLT~\cite{carlevaris2016university} is notable for its collection of campus-scale trajectories that span between roads and buildings. However, it exhibits several limitations. The elevation differences are relatively modest, and building traversal is often partial or discontinuous. Indoor segments are typically isolated and do not involve connected multi-floor navigation or long-range structural transitions within buildings.

Recent models such as SG-LPR~\cite{jiang2024sg} and IS-CAT~\cite{joo2024cat} have been evaluated on NCLT, but their experiments generally focus on localized segments or lightly mixed domains without explicitly addressing strong spatial or semantic transitions. For example, they often do not include continuous movement from enclosed corridors to open outdoor areas, where perceptual aliasing and geometric ambiguity can lead to degraded recognition performance. Although RoboLoc does not provide explicit annotations for such domain boundaries, its sequences inherently include a wide variety of environmental changes due to continuous recording across buildings, paths, and open regions. This structure allows for a more natural evaluation of algorithmic robustness in complex deployment settings.

This gap in existing resources highlights the need for a unified dataset that provides continuous LiDAR-based trajectories across both indoor and outdoor settings. Such a dataset should include topographic variation, a variety of road and corridor structures, and naturally occurring sensor degradation. RoboLoc addresses this need by capturing densely structured scenes that span multiple buildings, terrain levels, and environmental types within a single campus. It provides a realistic and comprehensive benchmark for evaluating generalization and robustness in modern 3D place recognition systems.

\section{RoboLoc: Indoor-Outdoor Benchmark Dataset}\label{sec4}

\subsection{Platform and Sensor Setup}

\begin{figure}[!t]
\centering
\includegraphics[width=0.4\textwidth]{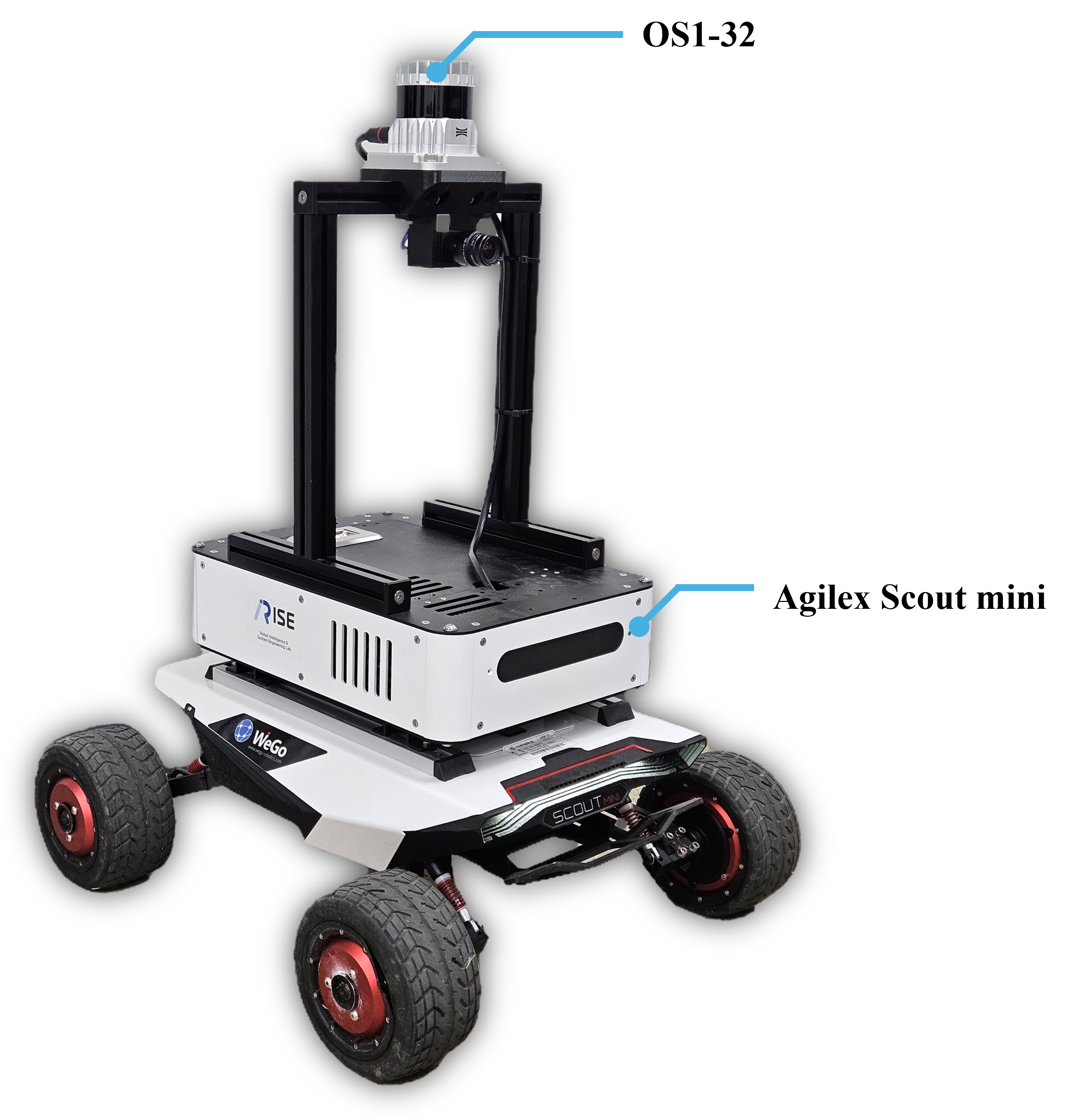}
\caption{The mobile data collection platform used in the RoboLoc dataset consists of an AgileX Scout Mini robot base equipped with an OS1-32 LiDAR sensor. This platform was deployed to navigate various indoor and outdoor areas across the Ajou University campus, serving as the primary system for collecting spatial data used to generate the RoboLoc maps.}
\label{fig:platform}
\end{figure}

The main objective of our dataset is to facilitate research in range sensor-based place recognition across a wide range of environments within a university campus, including areas where large survey vehicles cannot operate, such as narrow sidewalks and indoor corridors. To support data collection in such spaces, we utilize the \textbf{AGILEX SCOUT MINI}, a compact and agile four-wheeled mobile robot platform commonly used in robotics research and real-world field deployments. Its small form factor and high mobility make it well-suited for navigating both indoor and outdoor pedestrian-scale environments. An overview of the platform and its sensor mounting configuration is shown in Fig.~\ref{fig:platform}.

This platform is equipped with a single 3D LiDAR sensor, adhering to a minimalistic sensor configuration. This design reflects our goal of enabling practical and broadly deployable systems, particularly in GPS-denied environments where visual or inertial sensors may be unreliable due to lighting variations, magnetic interference, or limited mounting space.

We employ the \textbf{Ouster OS1-32} LiDAR, a 32-beam rotating 3D sensor known for its high resolution and long-range capabilities. It provides a full 360$^\circ$ horizontal field of view and operates at a frequency of 10 Hz, enabling the robot to continuously perceive its surrounding environment in real time. The LiDAR is securely mounted on top of the SCOUT MINI, at a height that minimizes occlusion from the robot body while maximizing visibility in all directions.

The sensor produces dense 3D point clouds in the form of $(x, y, z, \text{intensity})$ data. Each scan covers a wide vertical field of view, making it suitable for capturing indoor features such as walls, doorways, and ceilings, as well as outdoor features including building facades, vehicles, and vegetation. The effective sensing range extends up to 120 meters in optimal conditions, allowing robust detection of both nearby and distant landmarks.

All LiDAR frames are precisely timestamped and recorded at consistent intervals, ensuring temporal synchronization throughout the dataset. This temporal consistency supports various downstream tasks such as sequence matching, loop closure detection, and long-term localization.

\subsection{Target Environments and Sequence Detail}

\begin{figure}[!t]
\centering
\includegraphics[width=0.45\textwidth]{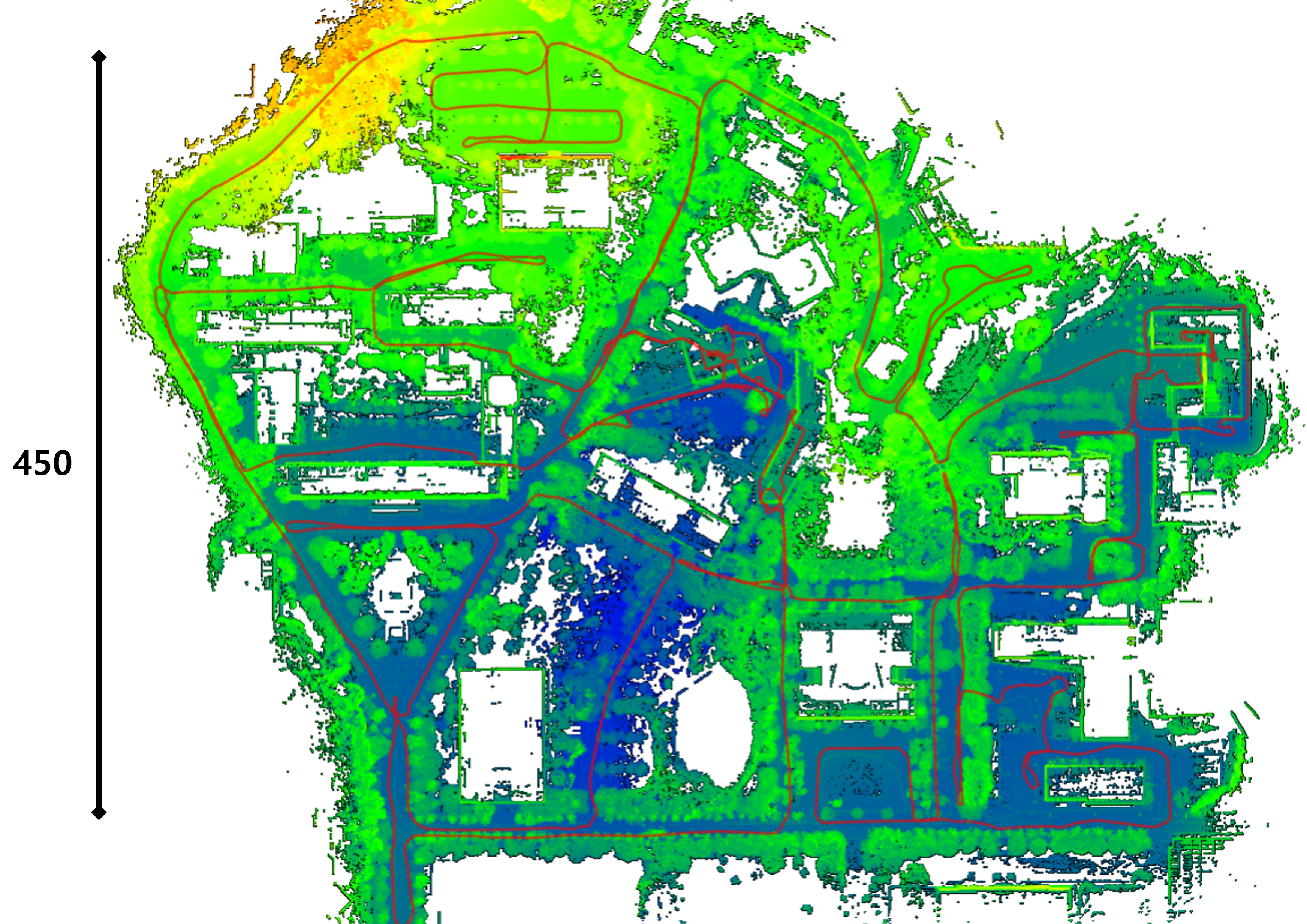}
\caption{A visualization showing the actual environment mapped using LiDAR data, along with the corresponding robot trajectory, overlaid on a high-resolution satellite image of Ajou University. The LiDAR map captures the 3D structure of the campus, while the trajectory is illustrated as a simplified linear path to provide a clear overview of the robot’s movement throughout various indoor and outdoor areas.}
\label{fig:map_full}
\end{figure}

The RoboLoc dataset was collected within the Ajou University campus, which offers a compact yet topographically and structurally diverse environment ideal for evaluating real-world place recognition systems. Unlike prior datasets that focus exclusively on either indoor or outdoor scenes, RoboLoc integrates both domains in a unified spatial map, capturing seamless transitions between roads, building entrances, and indoor corridors within a single recording sequence. This continuity reflects realistic deployment scenarios where mobile robots must handle mixed environments during long-range navigation tasks. An overview of the full mapped area and the robot's trajectories is shown in Fig.~\ref{fig:map_full}.

\begin{figure*}[t]
    \centering
    \includegraphics[width=0.9\textwidth]{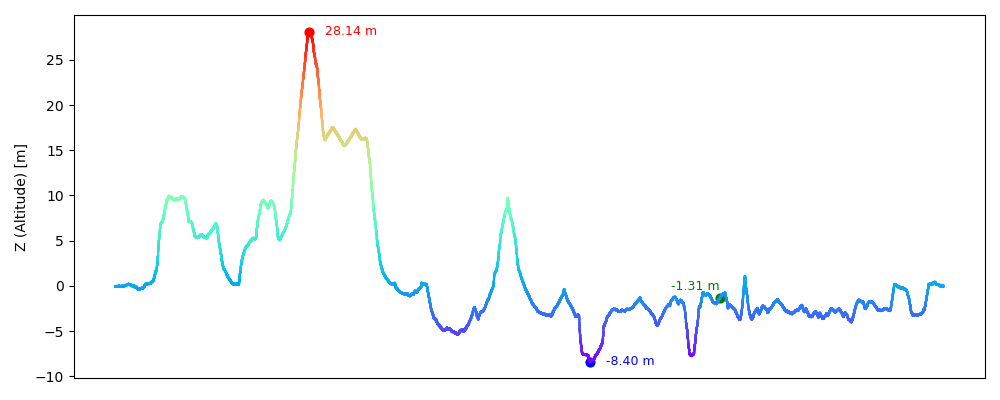}
    \caption{Visualization of altitude variation across the dataset. Red, blue, and green markers denote the maximum, minimum, and median elevations, respectively.}
    \label{fig:z_range}
\end{figure*}

Despite the relatively small overall size of the mapped area (approximately 450\,m $\times$ 600\,m), the campus presents significant topographic complexity. The elevation difference between the lowest and highest accessible points reaches approximately 36\,m, which can introduce considerable changes in viewpoint and geometric appearance. However, the median elevation variation within individual sequences remains modest at 1.31\,m, suggesting smooth transitions interspersed with steeper segments. An example of per-sequence elevation change is illustrated in Fig.~\ref{fig:z_range}.

Furthermore, RoboLoc includes navigation through \textbf{multi-floor indoor environments}, which pose a rarely addressed challenge in the context of LiDAR-based place recognition. Some buildings support continuous traversal from upper levels to ground floors via internal staircases, while others exploit terrain elevation to provide outdoor access at multiple floors. These multi-floor connections yield natural vertical transitions that mirror deployment scenarios in dense campus or building complexes.

\begin{figure*}[!t]
\centering
\includegraphics[width=1\textwidth]{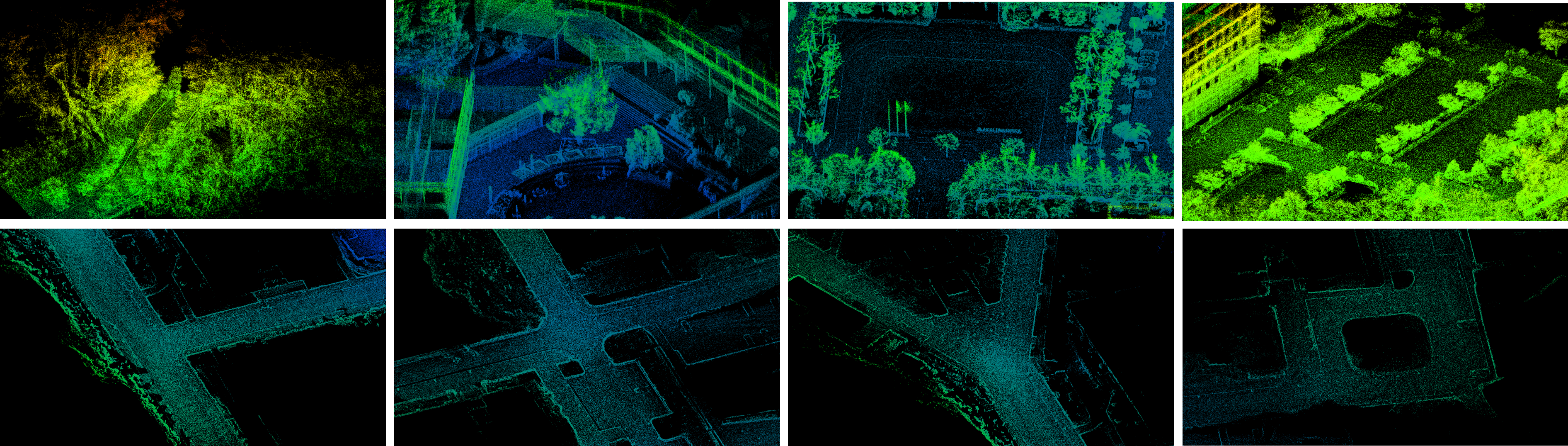}
\caption{Representative LiDAR maps illustrating the environmental diversity of our data collection sites. The first row includes (from left to right): a forest trail, a densely structured outdoor facility, a plaza area, and a parking lot. The second row showcases a variety of road scenes, including intersections and building-surrounded streets. This diversity in natural, architectural, and road environments provides challenging and realistic settings for evaluating place recognition and localization algorithms.}
\label{fig:str_diversity}
\end{figure*}

\begin{figure*}[!t]
\centering
\includegraphics[width=1\textwidth]{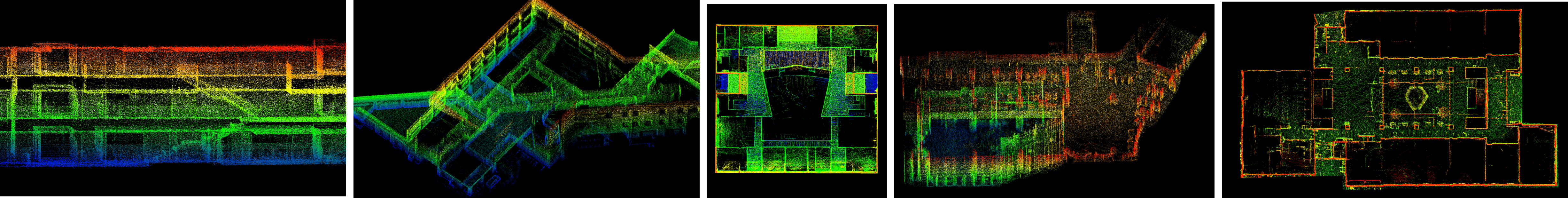}
\caption{Examples of challenging indoor scenarios captured in our dataset. The two leftmost scenes represent multi-floor buildings, characterized by vertically stacked structures and occluded corridors. The two rightmost scenes showcase large-scale, complex indoor environments with irregular geometries and cluttered spaces, presenting significant difficulties for 3D place recognition.}
\label{fig:indoor}
\end{figure*}

RoboLoc also exhibits a high degree of structural diversity in both outdoor and indoor environments. The road network comprises narrow pedestrian walkways, wide intersections, curved driveways, and building-adjacent alleys. Indoors, the robot encounters straight corridors, T-junctions, L-bends, and nested loops, which are architectural patterns that create perceptually similar yet topologically distinct scenes. Representative examples of these indoor environments are visualized in Fig.~\ref{fig:indoor}, while structural patterns are summarized in Fig.~\ref{fig:str_diversity}.

Beyond spatial layout, RoboLoc introduces several \textbf{perception-level challenges}. Indoors, narrow passages and reflective surfaces such as glass walls cause severe occlusions and LiDAR range dropouts. Outdoors, dense vegetation or repetitive architectural motifs lead to geometric ambiguities. Additionally, multiple locations within the dataset share highly similar structural layouts, especially those that are located within the same building or situated on different floors. This similarity induces strong perceptual aliasing. Repeated corridor widths, symmetric entrances, and regularly spaced junctions make disambiguation more difficult and increase the likelihood of false matches.

\begin{figure}[h]
\centering
\includegraphics[width=0.48\textwidth]{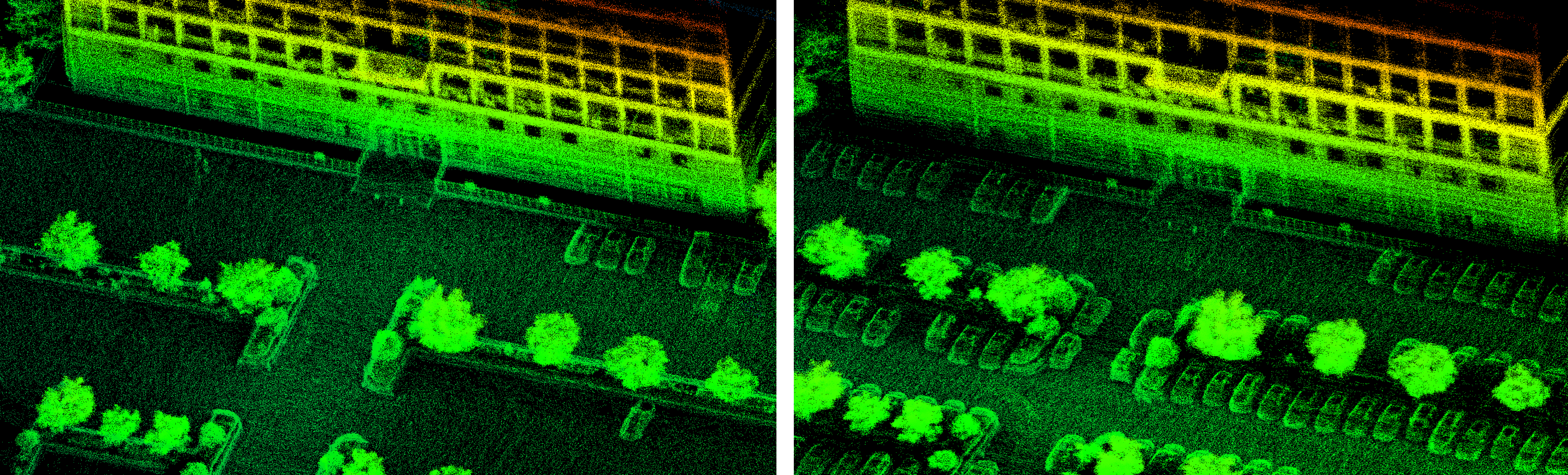}
\caption{Comparison of LiDAR maps recorded on a weekend (left) and a weekday (right) in the same parking area. Noticeable differences in the presence of parked vehicles highlight the temporal variation of static structures. These differences present challenges for consistent place recognition, emphasizing the need for temporal robustness.}
\label{fig:time_diversity}
\end{figure}

To incorporate \textbf{temporal diversity}, each location was recorded under two contrasting conditions: (1) early morning and weekend sessions with minimal dynamic activity, and (2) weekday daytime sessions with significant motion from pedestrians, bicycles, and occasional vehicles. This dual-mode acquisition captures environmental variability without requiring scene relabeling or simulation, and allows researchers to evaluate temporal robustness in localization and re-identification tasks. A visual comparison of these dynamic differences is shown in Fig.~\ref{fig:time_diversity}.

Overall, RoboLoc provides a challenging yet realistic testbed that captures the nuanced difficulties of spatial, structural, and perceptual variation in real-world navigation settings.

\subsection{Environmental Diversity and Transition Types}

A further strength of RoboLoc lies in its ability to naturally represent a wide spectrum of environments and transition types within single, unbroken trajectories. Rather than relying on manually segmented scenes, the dataset continuously records navigation across semantically distinct and geometrically diverse areas, such as indoor corridors, building exits, outdoor paths, open plazas, and forested terrain. This organic composition reflects the way mobile robots encounter environments in practice, without artificial boundaries or scripted changes.

The transition types embedded within RoboLoc include: (1) indoor-to-outdoor passages via doors or stairwells, (2) vertical level shifts enabled by terrain or floor connectivity, (3) spatial changes from open to narrow regions (e.g., plazas to hallways), and (4) semantic domain shifts between natural (e.g., vegetated) and built-up (e.g., sidewalk) scenes. These shifts are interleaved naturally within each trajectory, offering varied perceptual and geometric contexts without requiring scene-level annotations.

Such transitions compound the difficulty of place recognition by introducing abrupt changes in viewpoint, occlusion, and scene semantics. For instance, exiting a building can rapidly change the field of view and lighting conditions, while traversing vegetated areas introduces clutter, range dropouts, and ambiguous geometry. Moreover, environments with mirrored layouts or high structural regularity such as multi-floor interiors heighten the risk of perceptual aliasing. These characteristics make RoboLoc well suited for evaluating generalization under domain shifts and degraded sensing conditions.

Additionally, RoboLoc supports trajectory-based evaluation across domains, where algorithmic robustness can be tested not only on isolated keyframes but over extended navigation sequences involving cumulative environmental transitions. This property encourages development of systems that adapt over time and across scenes, mimicking real-world operational demands.

In summary, the embedded diversity of transition types and spatial semantics, combined with RoboLoc's seamless recording methodology, makes it an effective benchmark for studying both place-specific distinctiveness and domain adaptation in 3D perception systems. This diversity is further complemented by RoboLoc’s consistent sensor setup and synchronized temporal acquisition, enabling fair comparisons across scenarios. By encompassing both high-frequency transitions and long-range continuity, the dataset bridges the gap between controlled benchmarks and in-the-wild deployment, providing an essential resource for evaluating spatial understanding, robustness, and lifelong localization capabilities.

\subsection{Submap Generation}

To generate submaps from the point cloud data, we apply a uniform temporal sampling strategy along each recorded trajectory. In practice, we extract one submap every 2 seconds using the robot’s global pose, yielding roughly 105,514 submaps across the entire dataset.

Each submap contains 3D LiDAR points collected within a fixed spatial region centered at the robot’s position at the sampling time. This region is defined as a sphere with a 30-meter radius, or 60 meters in diameter, capturing the surrounding geometry near the robot. Alongside the point cloud, we also record the corresponding 6-DoF global pose (i.e., position and orientation) of the LiDAR sensor for each submap.

To ensure submap independence, which is particularly important for intra-sequence place recognition, we include only the LiDAR points captured within a one-second window around each sampling time. This avoids overlaps between adjacent submaps and maintains both spatial and temporal consistency. Overall, this submap generation process provides a structured and reliable setup for training and evaluating models in both intra- and inter-sequence retrieval scenarios.

\subsection{Trajectory Generation for Ground Truth}
Accurate and globally consistent trajectories are essential for constructing a reliable place recognition benchmark. While GNSS or motion capture systems are typically used for trajectory ground truth in outdoor settings, such infrastructure is either unavailable or impractical in indoor environments. Given that our dataset spans both indoor and outdoor scenes, we require a unified, sensor-based approach that does not depend on external localization systems.

To this end, we adopt \textbf{GLIM} (GPU-accelerated LiDAR-Inertial Mapping)~\cite{koide2024glim}, a state-of-the-art SLAM framework that performs real-time 3D mapping through tight LiDAR-IMU integration. GLIM leverages fixed-lag smoothing and keyframe-based LiDAR scan matching within a sliding optimization window, allowing for stable odometry even in geometrically degenerate environments such as long corridors or areas with sparse structural features. Unlike filtering-based methods that instantly commit to state estimates, GLIM continuously refines past states, yielding globally consistent trajectories.

In our dataset, we deliberately run GLIM in LiDAR-only mode, excluding visual and GNSS inputs to ensure fair evaluation in GPS-denied and lighting-variable environments. This configuration aligns with our focus on LiDAR-centric place recognition and supports consistent trajectory generation across both indoor and outdoor segments of the dataset.

\begin{figure}[h]
\centering
\includegraphics[width=0.45\textwidth]{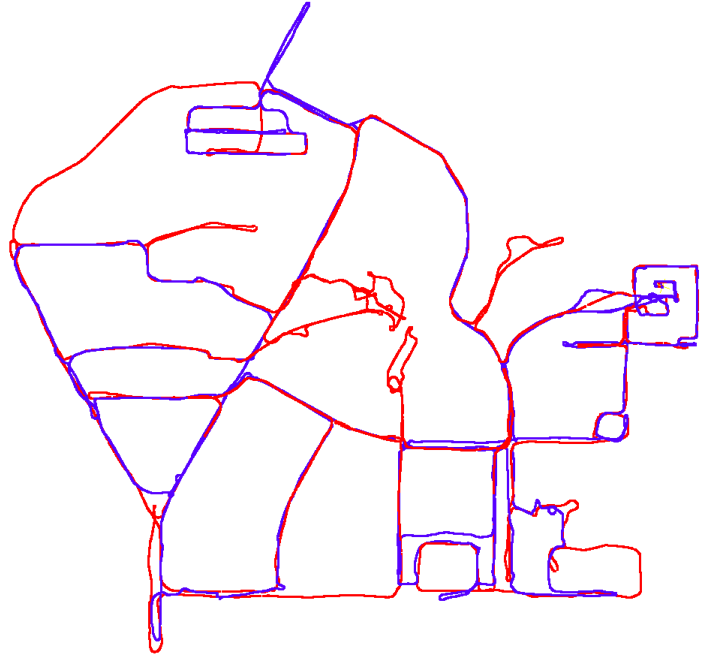}
\caption{Overlay of two GLIM-generated trajectories recorded at different times. The dominant magenta color indicates strong spatial consistency, as the red and blue paths nearly coincide throughout the entire route.}
\label{fig:glim_consistency}
\FloatBarrier
\end{figure}

To empirically verify the consistency of the resulting trajectories, we collected multiple repeated traversals over shared locations under different conditions. As illustrated in Fig.~\ref{fig:glim_consistency}, we overlay two trajectories from separate recordings performed at different times of day. The resulting visualization shows that the paths exhibit near-complete spatial overlap, with most deviations remaining within a sub-meter range. The dominant magenta coloring resulting from the overlap of red and blue trajectories visually confirms strong repeatability. \jjj{These results empirically demonstrate that accumulated drift is negligible for the purpose of place recognition benchmarking. Therefore, the GLIM-generated trajectories can be regarded as sufficiently reliable pseudo ground truth, even in the absence of GNSS or motion capture systems, particularly for evaluating models in GPS-denied and structurally complex environments.}

\subsection{Train/Test Splits and Evaluation Protocol}

Given the hybrid nature of the RoboLoc dataset, which contains both indoor and outdoor navigation scenarios, we adopt a modified evaluation protocol inspired by the Oxford RobotCar benchmark but with stricter and more spatially-sensitive constraints tailored for high-resolution LiDAR-based place recognition.

Each recorded sequence is temporally partitioned into mutually exclusive \textit{reference} and \textit{query} sets, ensuring there is no temporal overlap or information leakage between the sets. To construct these sets, submaps are uniformly sampled along the trajectory at fixed intervals of 0.5 seconds.

Positive matches are defined using a strict spatial proximity criterion: we consider a reference-query pair as a match if their ground-truth locations are within a Euclidean distance of \textbf{5 meters}. This tighter bound reflects the higher spatial precision needed for indoor scenarios, where place ambiguity is more common.

Negative matches are defined as reference-query pairs that are separated by more than \textbf{25 meters} in Euclidean distance. Reference submaps that lie between 5 and 25 meters from a query are excluded from evaluation to avoid ambiguous supervision during training and testing.

To avoid trivial retrievals, we additionally exclude all reference submaps that are within \textbf{7 meters} of the query along the same trajectory. This ensures that the model must retrieve scenes that are not simply adjacent in time or space, thereby enforcing meaningful recognition of spatially distinct locations.

This protocol allows for robust evaluation of both \textbf{intra-sequence} and \textbf{inter-sequence} retrieval tasks. Intra-sequence evaluation includes bidirectional revisits such as forward vs. reverse traversal of the same route, while inter-sequence evaluation tests robustness across recordings made at different times of day, under varying dynamic conditions (e.g., static vs. crowded environments), and in diverse lighting settings.

To quantify performance, we report standard retrieval metrics: \textbf{Recall@1}, \textbf{Recall@5} and \textbf{Recall@1\%}. These metrics are computed over all query submaps for each evaluation scenario, providing a comprehensive picture of the model’s precision, recall sensitivity, and ranking quality. In future extensions of the dataset, additional evaluation metrics such as localization error or temporal consistency scores could also be integrated to support broader benchmarking efforts.

\subsection{Data Structure and Format}

The RoboLoc dataset is organized into multiple sequence-level folders, each representing a complete data capture session under a specific environmental setting (e.g., indoor, outdoor, or mixed). These sequences are stored directly under the root directory of the dataset, with a consistent folder structure to facilitate parsing and integration into place recognition pipelines.

Each data sequence in the RoboLoc dataset is stored in its own directory (e.g., \texttt{sequence\_01/}, \texttt{sequence\_02/}), and contains two main components:

\begin{itemize}
  \item A \texttt{points/} folder, which contains timestamped LiDAR scans in `.bin` format. Each file represents one full sweep captured by the Ouster OS1-32 sensor at 10 Hz, and includes $(x, y, z)$ coordinates for all valid points.
  \item A \texttt{pose.txt} file, which records the ground-truth 6-DoF pose of the robot at each timestamp. Each line follows the format: \texttt{timestamp x y z qx qy qz qw}, where the position is given in meters and orientation as a quaternion.
\end{itemize}

This per-sequence organization ensures modularity and enables independent use of each sequence for training or evaluation. Unlike datasets that intermix all data globally, RoboLoc keeps sequence boundaries intact to avoid data leakage and support clean train-test splits.

\begin{table}[h]
\centering
\footnotesize
\caption{Distribution of Environmental Types in RoboLoc}
\label{tab:env_ratio}
\begin{tabular}{l c l}
\toprule
\textbf{Type} & \textbf{Coverage (\%)} & \textbf{Examples} \\
\midrule
Indoor & 25 & Corridors, entrances \\
Sidewalk / Road & 47 & Paved paths, vehicle lanes \\
Campus Open Space & 18 & Plazas, courtyards \\
Vegetated / Forested & 10 & Trees, grass areas \\
\bottomrule
\end{tabular}
\end{table}

\section{Evaluation and Benchmarking}\label{sec5}

\subsection{Baselines and Settings}

To evaluate the effectiveness of the proposed RoboLoc dataset, we benchmark three representative learning-based 3D place recognition models: PointNetVLAD, MinkLoc3D, and PTC-Net. These models are selected based on their wide adoption in prior literature and their distinct architectural philosophies, offering a meaningful contrast in feature representation and computational strategy.

\textbf{Point-based:} \textit{PointNetVLAD}~\cite{uy2018pointnetvlad} combines local geometric encoding via PointNet with global descriptor aggregation using NetVLAD pooling. Its design enables efficient processing of unordered point sets and yields compact descriptors that are well-suited for large-scale retrieval tasks. Due to its relatively lightweight nature and low computational overhead, it is also practical for real-time or embedded applications.

\textbf{Voxel-based:} \textit{MinkLoc3D}~\cite{komorowski2021minkloc3d} operates on voxelized representations of point clouds and leverages sparse 3D convolutions to learn hierarchical spatial features. Its architecture is particularly effective at handling variable point densities and spatial irregularities, which are common in real-world environments such as indoor corridors, open plazas, and vegetated paths. This makes it a strong candidate for robust place recognition under diverse environmental conditions.

\jjj{\textbf{Transformer-based:} \textit{PTC-Net}~\cite{chen2023ptc} adopts a point transformer convolutional backbone that unifies local geometric modeling and long-range feature interactions. By integrating transformer layers into the 3D recognition pipeline, PTC-Net and its lightweight variant (PTC-Net-L) aim to improve generalization under complex structural repetition and domain transitions. These models represent a new direction in point cloud retrieval by balancing expressiveness with scalability.
}

\jjj{All} models are evaluated under consistent experimental settings. We apply the same submap generation policy, temporal sampling intervals, and reference/query splitting strategy as described in Section~3. To ensure a fair comparison, we also maintain identical positive/negative matching constraints across methods.

Retrieval performance is measured using three standard metrics. Recall@1 indicates the percentage of queries for which the top-ranked database submap is a correct match. Recall@5 extends this by checking whether a correct match appears within the top five retrieved candidates, capturing near-miss retrieval accuracy. Recall@1\% evaluates whether the correct place appears within the top 1\% of the ranked database, offering a broader view of retrieval robustness under uncertainty. Together, these metrics provide a comprehensive assessment of both precision and tolerance in place recognition.

All implementations are based on publicly available codebases, and we preserve the original training hyperparameters, including batch size, learning rate, and augmentation policy. This reproducible evaluation pipeline allows us to fairly assess each model’s generalization ability on the RoboLoc dataset.

Overall, this benchmarking setup offers a strong foundation for understanding how different neural architectures respond to the complex spatial layout, structural repetition, and domain transitions present in RoboLoc’s real-world navigation scenarios.

\subsection{Quantitative Results}

\jjj{Table~\ref{tab:quant_results} summarizes the retrieval performance of learning-based place recognition models on the RoboLoc dataset, evaluated using Recall@1, Recall@5, and Recall@1\%. Among the evaluated methods, PointNetVLAD achieved the highest Recall@1\% of 61.0\%, followed by MinkLoc3DV2 and MinkLoc3D. Interestingly, MinkLoc3D and MinkLoc3DV2 showed identical Recall@1 (43.8\%), but MinkLoc3DV2 slightly outperformed at higher thresholds (e.g., Recall@5 and Recall@1\%). In contrast, PTC-NET and its light variant (PTC-NET-L) underperformed across all metrics, likely due to their original focus on urban outdoor environments. These results highlight the domain-adaptive challenges of mixed indoor–outdoor settings and the need for more robust and generalizable place recognition models.}

\begin{table}[h]
\centering

\caption{Retrieval performance of PointNetVLAD, MinkLoc3D, and \jjj{PTC-Net}\\ on the RoboLoc dataset, reported in terms of Recall@1, Recall@5,\\ and Recall@1\%.}
\label{tab:quant_results}
\resizebox{0.95\columnwidth}{!}{
\begin{tabular}{lccc}
    \toprule
    \textbf{Method} & \textbf{Recall@1 (\%)} & \textbf{Recall@5 (\%)} & \textbf{Recall@1\% (\%)} \\
    \midrule
    PointNetVLAD \cite{uy2018pointnetvlad}  & 43.2 & 47.3 & 61.0 \\ 
    MinkLoc3D \cite{komorowski2021minkloc3d} & 43.8 & 46.6 & 48.5 \\
    MinkLoc3DV2 \cite{komorowski2022improving} & 43.8 & 47.7 & 49.1 \\
    PTC-NET \cite{chen2023ptc} & 38.2 & 42 & 45.5 \\
    PTC-NET-L \cite{chen2023ptc} & 38.1 & 41.4 & 45 \\
    \bottomrule 
\end{tabular}
}
\end{table}
\begin{figure}[h]
    \centering
    
    \includegraphics[width=0.48\textwidth]{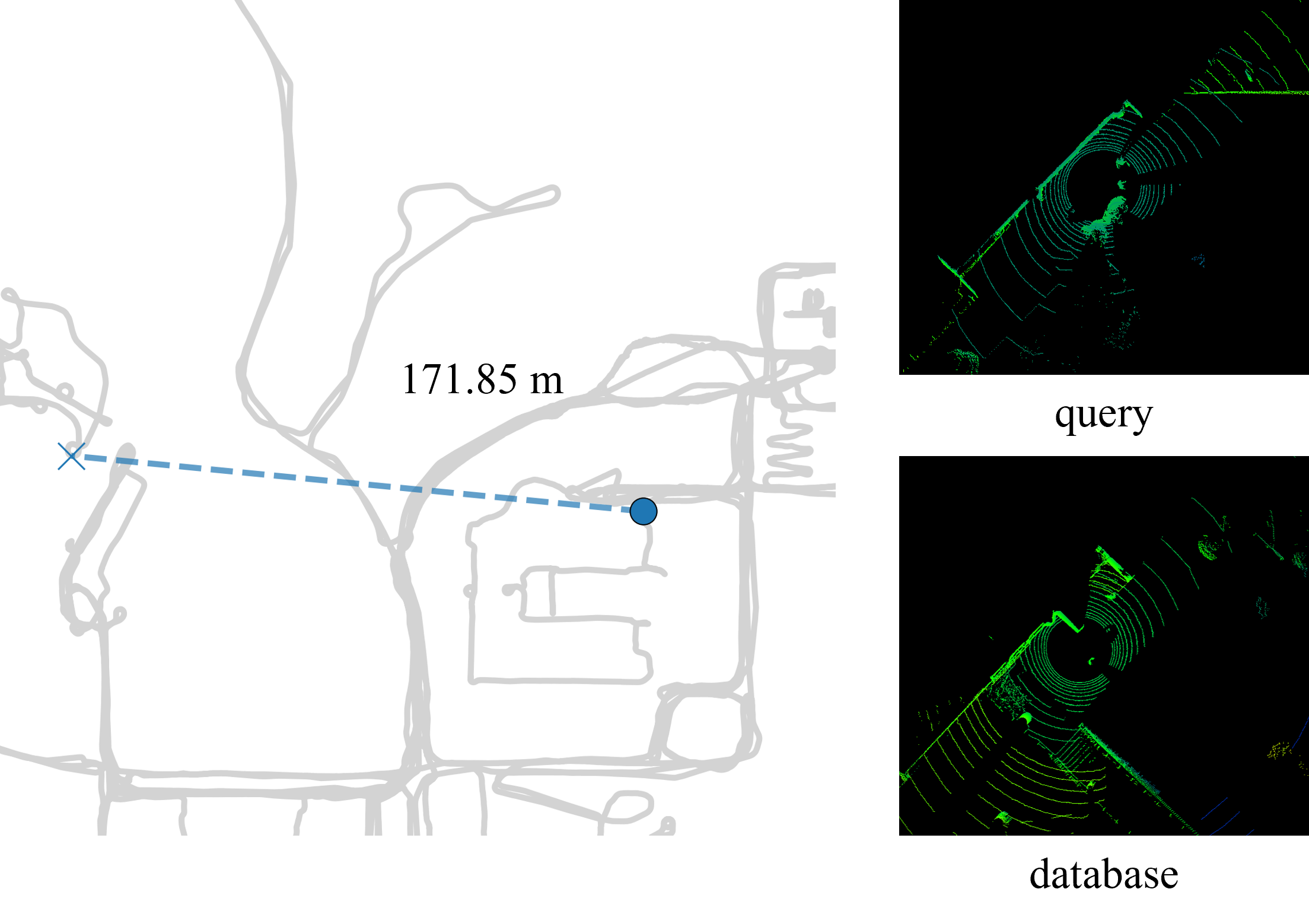}    
    \caption{Visualization of an unmatched query and database pair. The left shows the top-down map with the query location (blue circle) and its incorrect nearest database candidate (blue X), which are 171.85 m apart. The right side presents the corresponding point clouds: top is the query, and bottom is the retrieved database. Despite structural similarity, the model failed to retrieve the correct location due to large spatial separation and appearance variation.}
    \label{fig:query}
\end{figure}

\subsection{Qualitative Analysis} 
\jjj{
To further interpret the retrieval performance, we conducted a qualitative analysis of representative queries across indoor, outdoor, and transitional regions of the RoboLoc dataset.}
\jjj{
In indoor environments with repetitive structures such as corridors and stairwells, \textbf{MinkLoc3D} consistently produced accurate top-1 matches, demonstrating strong robustness to visual aliasing. In contrast, \textbf{PointNetVLAD} often retrieved incorrect locations that appeared structurally similar, likely due to its limited global context modeling and sensitivity to local feature repetition.}

\jjj{
In outdoor and semi-structured scenes, both MinkLoc3D and PointNetVLAD performed reasonably well. However, MinkLoc3D exhibited better stability under partial occlusion and viewpoint changes, particularly in domain transition scenarios, e.g., exiting buildings into open courtyards. This highlights the advantage of hierarchical and voxel-based representations in ambiguous or mixed-domain environments.
We also evaluated \textbf{PTC-Net} and \textbf{PTC-Net-L}, which integrate transformer-based modules. Although these models leverage point-wise geometric relations, they showed reduced robustness in indoor scenes, especially at domain transition points. We observed increased failure rates when encountering reflective surfaces or significant appearance changes. This suggests that transformer-based methods may require additional adaptation mechanisms to generalize across highly diverse domains.}

\jjj{
As illustrated in Fig.~\ref{fig:query}, \textbf{PointNetVLAD} exhibited a failure case at a building exit with large glass doors. The model incorrectly retrieved a visually similar but geographically distant indoor submap due to strong reflections and structural symmetry. This highlights the vulnerability of local-feature-based methods to reflective surfaces. Similar issues were occasionally observed in other models as well, underscoring the inherent challenge of domain transitions in complex environments.
These examples underline the importance of developing reflection-robust features and domain-invariant representations for future models.}


\subsection{Failure Cases and Limitations}
Despite promising results, both models exhibit consistent failure cases, especially in structurally repetitive or visually ambiguous regions.

A common failure mode was perceptual aliasing in indoor corridors, where repeated geometries led to false positive matches. This issue was more frequently observed in PointNetVLAD, likely due to its less expressive global descriptors. MinkLoc3D demonstrated better robustness in such cases but was not completely immune to errors.

Another challenge emerged in vegetated or semi-structured outdoor areas, where LiDAR reflections were sparse or noisy. Although not yet benchmarked in this study, prior work suggests that rule-based methods such as Scan Context or ICP tend to perform poorly in these conditions due to limited robustness to occlusion and structural variability.


\section{Future Work}

\jjj{Future versions of RoboLoc will incorporate point-level semantic annotations, per-point intensity values, and multi-floor indoor trajectories to enable more comprehensive benchmarking. We also plan to include additional sequences under diverse lighting and seasonal conditions to support evaluations of temporal robustness and long-term generalization. Through these extensions, RoboLoc aims to evolve into a unified benchmark for robust, geometry-centric localization across spatial, semantic, and temporal domains.}

\jjj{Furthermore, we envision RoboLoc being extended beyond place recognition to support broader research directions. Its structural diversity and domain transitions make it a promising platform for evaluating transformer-based architectures~\cite{duan2023dynamic}, parameter-efficient fine-tuning methods~\cite{duan2025parameter}, and semi-supervised learning frameworks~\cite{duan2024wearable}. With future integration of semantic labels, RoboLoc could also be adapted for tasks involving language-guided 3D understanding~\cite{miao2025laser} and lifelong learning in evolving environments~\cite{duan2024dual}.}

\section{Conclusion}

We introduced \textbf{RoboLoc}, a LiDAR-based dataset designed for place recognition across both indoor and outdoor environments. It captures continuous transitions between structurally and semantically diverse areas, including hallways, campus roads, vegetated paths, and open plazas, thereby supporting unified evaluation across domains within a single sequence.

To demonstrate its effectiveness, we benchmarked two representative models: PointNetVLAD, MinkLoc3D, and PTC-Net. The results revealed complementary strengths depending on scene characteristics, highlighting the importance of architectural diversity when addressing spatial variation and perceptual ambiguity.

RoboLoc serves as a realistic testbed for developing and evaluating place recognition methods under varied conditions.

\section*{Acknowledgment}
This work is supported by the Korea Agency for Infrastructure Technology Advancement(KAIA) grant funded by the Ministry of Land,Infrastructure and Transport (Grant 1615013176)

\bibliographystyle{ieeetr}
\bibliography{paper}
\end{document}